\documentclass[conference]{IEEEtran}
\IEEEoverridecommandlockouts
\usepackage{cite}
\usepackage{amsmath,amssymb,amsfonts}
\usepackage{algorithmic}
\usepackage{graphicx}
\usepackage{textcomp}
\usepackage{xcolor}
\usepackage{hyperref}
\usepackage{graphicx}
\usepackage{latexsym}
\usepackage{booktabs}
\usepackage{diagbox} 
\usepackage[ruled, noend]{algorithm2e}
\usepackage{multirow}

\def\BibTeX{{\rm B\kern-.05em{\sc i\kern-.025em b}\kern-.08em
    T\kern-.1667em\lower.7ex\hbox{E}\kern-.125emX}}
\begin{document}

\title{Collaborative Batch Size Optimization for \\ Federated Learning
\thanks{This work was funded in part by the Luxembourg
National Research Fund (FNR) under grant number 18047633.}
}

\author{\IEEEauthorblockN{Arno Geimer, Karthick Panner Selvam, Beltran Fiz Pontiveros}
\IEEEauthorblockN{Interdisciplinary Centre for Security, Reliability and Trust\\
University of Luxembourg\\
\{arno.geimer, karthick.pannerselvam, beltran.fiz\}@uni.lu,}}

\maketitle

\begin{abstract}

Federated Learning (FL) is a decentralized collaborative Machine Learning framework for training models without collecting data in a centralized location. It has seen application across various disciplines, from helping medical diagnoses in hospitals to detecting fraud in financial transactions. In this paper, we focus on improving the local training process through hardware usage optimization. While participants in a federation might share the hardware they are training on, since there is no information exchange between them, their training process can be hindered by an improper training configuration. Taking advantage of the parallel processing inherent to Federated Learning, we use a greedy randomized search to optimize local batch sizes for the best training settings across all participants. Our results show that against default parameter settings, our method improves convergence speed while staying nearly on par with the case where local parameters are optimized.

\end{abstract}


\section{Introduction}

Following the introduction of data privacy regulations such as GDPR or CCPA, efforts have been made to allow the usage of private data sources while respecting data privacy laws. Federated Learning \cite{mcmahan2017communication}, a collaborative machine learning technique that preserves data privacy, has emerged as a prominent approach in related research. It allows for the training of a shared model between a group of data owners, without any data leaving a participant's premises. It has seen usage in different domains, including healthcare, finance\cite{li2020review}, and distributed Large Language Model training\cite{gao2025flowertune}.
As a privacy-focused decentralized Machine Learning paradigm, a lack of information sharing results in a sub-par outcome. This paper addresses the following challenge: Given a federation with fixed hardware resources, what is the maximum shared batch size that can be physically supported? Following previous work on the effectiveness of batch size changes in Federated Learning, we propose a novel method to determine boundary values related to client-side hardware. Our results show that with our method we can achieve close to optimal performance when compared to default values, without running the risk of overloading clients and a subsequent collapse of the federation.

The contributions of this paper are as follows.
\begin{enumerate}
    \item We introduce a Randomized Binary Search-Based Mechanism for determining the maximum achievable shared batch size in a federation.
\end{enumerate}

In the following sections, we detail the background and related work, introduce our method, and demonstrate its effectiveness in reducing simulation time.

\section{Background}

\subsection{Federated Learning}

Federated learning is a decentralized Machine Learning framework in which a group of data owners, called clients or participants, collaboratively train a shared model. A federation consists of participants and a central server that manages the synchronization, scheduling, and model update step. Each participant trains the model locally on their own data for some epochs, after which the central server calculates an aggregation of all local models. This aggregation is transmitted to all clients and is used as a starting point for further training. This allows each member of the federation to acquire a shared model that contains knowledge of all the other members' data, without any of the data having to leave the premises of its owner.

Over the years, multiple challenges have emerged in the deployment and training of Federated Machine Learning models. Persuading potential participants to engage in a federation is not trivial in non-research settings. Another problem are malicious actors, who, as research has shown, can fully recover the singular data points of other participants without proper countermeasures. An important challenge is the reluctance of participants to share information. In a centralized federation, where a central server manages the operation, or in a decentralized federation, where clients communicate directly between each other. Due to regulations or simply due to mistrust, participants do not share more information than they deem necessary. However, a certain transparency can be crucial for the federation. For example, in the case where clients' datasets are highly heterogeneous, some aggregation methods can yield better results than in a homogeneous federation. Additionally, with varying hardware capabilities, personalized scheduling can be beneficial in accelerating the FL process. Without knowing of these differences, the federation does not perform at its optimum.
The challenge we focus on in this paper is the central server's unawareness of the hardware capabilities of participants. We analyze the impact of and potential solutions to the use of a hardware-constrained parameter, namely batch size, in Federated Learning.

\subsection{Mini-Batch Training}\label{sec:mbt}

In Machine Learning, a mini-batch denotes the subset of training data used to calculate a gradient in a single step of backpropagation. This is in contrast to full-batch training, where the gradient is calculated each epoch on the full training data. After each batch iteration, the model parameters are updated accordingly. The amount of training data present in a batch, or batch size, varies depending on the type of data set, the size of the data points, and the hardware. As modern Machine Learning models are overwhelmingly trained on GPUs, batch sizes are an important parameter in the training configuration due to memory restrictions. During training, data is sent to the GPU as a batch, leading to a direct relation between GPU memory usage and batch size. Although some data sets may be small enough to be fully loaded into GPU memory (VRAM), larger datasets far exceed modern GPU memory capabilities. Hence, the need for splitting data into mini-batches. Although the total amount of data processed in an epoch does not change between different batch sizes, the number of memory load operations decreases with a larger batch size. Higher batch sizes therefore result in lower execution time. However, this does not mean that fully utilizing GPU memory through maximizing batch size is advisable.
Batch sizes have been shown to have a substantial influence on the stability and end result of the training process in Machine Learning. Although efforts have been made in the domain of large batch size training\cite{smith2017don}\cite{you2020limit}, small batch sizes have been shown to lead to more stable training and higher model accuracy\cite{he2019control}\cite{masters1804revisiting}. The trade-off between good convergence and lower computation time is an important part of hyperparameter optimization in Machine Learning. An additional variable in this problem is the restriction of batch size due to hardware constraints. In fact, the GPU might simply not have enough memory to fully load a batch, which can terminate the training process. 

\subsection{Randomized Binary Search}

Given an array of length $n$, search algorithms are used to find a specific element in the most optimal time. For example, a linear search iterates through an array until the corresponding element is found. It has a time complexity of $O(n)$, regardless of whether the array is sorted or not. Assuming the array is sorted, a binary search is more effective. Instead of iterating over every single element, the binary search splits the array into two parts and determines in which part the wanted element would lie. Iterating this process until the element is found, a binary search outperforms a linear search by decreasing the complexity to $O(\log(n))$. The same time complexity applies to a randomized binary search, where the array is split into two parts of random lengths. In this work, we will use a federated version of randomized binary search, taking advantage of the federation's multiple participants to further improve search performance.

\section{Related Work}

\subsection{Batch size in Federated Learning}\label{sec:bsrestr}

Although full batch training has been proposed in Federated Learning\cite{mcmahan2017communication}, federations usually use mini-batch-based training. This adds the minibatch size as an additional parameter to the federation and raises the problem of setting an appropriate value. While most work chooses to establish a shared minibatch size among all participants, the question of client-specific batch sizes in Federated Learning has been studied before. Adaptive batch sizes can be proposed for specific clients to deal with stragglers\cite{ma2021adaptive}\cite{liao2024mergesfl}\cite{liu2023dynamite}.Knowing the hardware capabilities and expected time to update of participants allows the adaptation of batch size, as well as local epochs, to minimize downtime. Other works use batch sizes specific to the client to reduce the overall communication cost\cite{liu2020adaptive}\cite{liu2023adacoopt} arising from model update communication with the central server. It has been suggested to gradually increase\cite{shi2022talk} batch size as a solution to the trade-off between communication and computation time. Although these previous papers analyzed hardware capabilities, they all choose initial batch size boundaries that fit their hardware\cite{park2022amble}. To the best of our knowledge, no previous work has proposed a mechanism that lets the federation itself determine if a proposed batch-size limit may be too large for some participants' hardware.

\section{Our method}

We use a greedy randomized binary search algorithm to determine optimal batch sizes in the first round(s) of the federation. Specifically, we initially send a minimum batch size, $b_{min}$, to all clients along with the initial model parameters. Each client in parallel $i$ randomly samples a batch size $b_i > b_{min}$, limited by the size of their local dataset. If training is successful using the chosen batch size, we can use $b_i$ as the new minimum batch size $b_{min}$. Otherwise, if training does not succeed, $b_i$ was too large and is thus an upper bound for the maximum batch size. In this case, we set a maximum batch size $b_{max} = b_i$. A detailed outline of our method can be found in Algorithm \ref{alg:rasba}. Since the model does not update if all clients sample too large batch sizes, we employ error handling to prevent the loss of a full round of training. Until a shared batch size has been determined, a fraction $f$ of participants do not participate in the search and instead train the model using $b_{min}$ as batch size. Due to a higher sampling rate compared to the generic randomized binary search, our method reduces the time to convergence by $((1 - f) \times m)^{-1}$ compared to the binary search, where $m$ is the number of participants.

\begin{algorithm}
\DontPrintSemicolon
\caption{Randomized binary search for batch optimization in a federation with $m$ clients.}
\label{alg:rasba}
\textbf{Server} sets initial model $m_t$, ($b_{min}, b_{max}$) \;
\For{\upshape each round $t = 1, 2, ...$}{
    \textbf{Server} sends $b_{min}, b_{max}$, $m_{t}$ to clients\;
    \For{\upshape each client $i \leq n$}{
        \If{$b_{min} \neq b_{max}$}{
            Randomly sample $b \in [b_{min}, b_{max}]$ \;
            \textbf{Try:} Update model $m_t^i$ using batch size $b$ \;
            \hspace{0.6cm} $b_{min}^i = b$ \;
            \textbf{Except} OutOfMemoryError: \;
            \hspace{0.6cm} $b_{max}^i = b$\;
        }
        \textbf{else} Train using batch size $b_{min} = b_{max}$\;
        \textbf{Send}  $m_t^i$, ($b_{min}, b_{max}$) to server \;
    }
    \textbf{Server} sets $b_{min} = \text{max}_i(b_{min}^i), b_{max} = \text{min}_i(b_{max}^i)$
}
\end{algorithm}

Naturally, the initial value for $b_{max}$ determines the speed with which our method finds an optimal shared mini-batch size. However, $b_{max}$ is inherently bounded by the minimum data set size of the participants: Clients cannot load more data in a batch than they have, yielding a maximal value for $b_{max}$. We further restrict $b_{max}$ following the observations in Section \ref{sec:mbt} and set $b_{max} = 64$ in the initial step. It is crucial to note that this is not equivalent to starting a federation with a set batch size of 64. If any of the clients cannot handle such a batch size, our method will naturally decrease the shared batch size until a common value is found.

\begin{table*}[!htbp]
    \begin{center}
    \begin{tabular}{cc|c|c||c|c|c}
        \cline{2-7}
        &\multicolumn{3}{c||}{\textsc{Mnist}}& \multicolumn{3}{c}{\textsc{Cifar10}}\\
        \cline{1-7}
        \multicolumn{1}{c||}{batch} & time (s) & speedup ($\times$) & acc. (\%) & time (s) & speedup ($\times$) & acc. (\%) \\
        \cline{1-7}
        \multicolumn{7}{c}{} \\
        \cline{1-7}
        \multicolumn{1}{c||}{4} & 5363 & 1 & 90.76 & 2879 & 1 & 67.43 \\
        \multicolumn{1}{c||}{8} & 2823 & 1.9 & 94.95 & 1683 & 1.71 & 72.41 \\
        \multicolumn{1}{c||}{16} & 1530 & 3.51 & 94.01 & 1083 & 2.66 & 73.95 \\
        \multicolumn{1}{c||}{32} & 885 & 6.06 & 93.09 & 820 & 3.51 & 73.66 \\
        \multicolumn{1}{c||}{64} & 559 & 9.6 & 91.21 & 607 & 4.74 & 70.62 \\
        \multicolumn{1}{c||}{128} & 416 & 12.89 & 87.84 & 936 & 3.08 & 68.05 \\
        \multicolumn{1}{c||}{256} & 342 & 15.68 & 83.17 & 494 & 5.83 & 64.67 \\
        \cline{1-7}
        \multicolumn{7}{c}{} \\
        \cline{1-7}
        \multicolumn{1}{c||}{our method} & 691 & 7.76 & 96.09 & 602 & 4.78 & 73.42 \\
        \cline{1-7}
        \multicolumn{7}{c}{} \\
       \end{tabular}
       
       \caption{Speed and performance of different batch sizes and our method.}
       \label{table:speed}
    \end{center}
\end{table*}

\section{Experimental results}

\subsection{Setup}

All code used in this paper is freely available on \href{https://anonymous.4open.science/r/batch-optimization-F567/}{Anonymous Github}. The Federated Learning framework used is Flower 1.18\cite{beutel2020flower} with PyTorch 2.7.0. A 1.6M parameter MobileNet-v3\cite{howard2017mobilenets} is trained on \textsc{Mnist}\cite{deng2012mnist}, while a 11.7M parameter ResNet18\cite{he2016deep} is trained on \textsc{Cifar10}\cite{krizhevsky2009learning}. The experiments were carried out on an Nvidia GeForce GTX 1080 with 8GB of VRAM, coupled with a 4-core Intel i3 8100 and 16 GB of RAM. Using an Adam optimizer, set an initial learning rate of $10^{-3}$ for all batch sizes. The data sets are split into 10 parts following a Dirichlet split\cite{wang2020federated} with $\alpha = 10$. Federations on \textsc{Mnist} run for 25, while federations on \textsc{Cifar10} run for 20 global rounds, both with an epoch size of 1. The aggregation strategy is set as \textsc{FedAvg}.

\subsection{Results}

\begin{figure}[h]
    \begin{center} 
    \includegraphics[scale = .375]{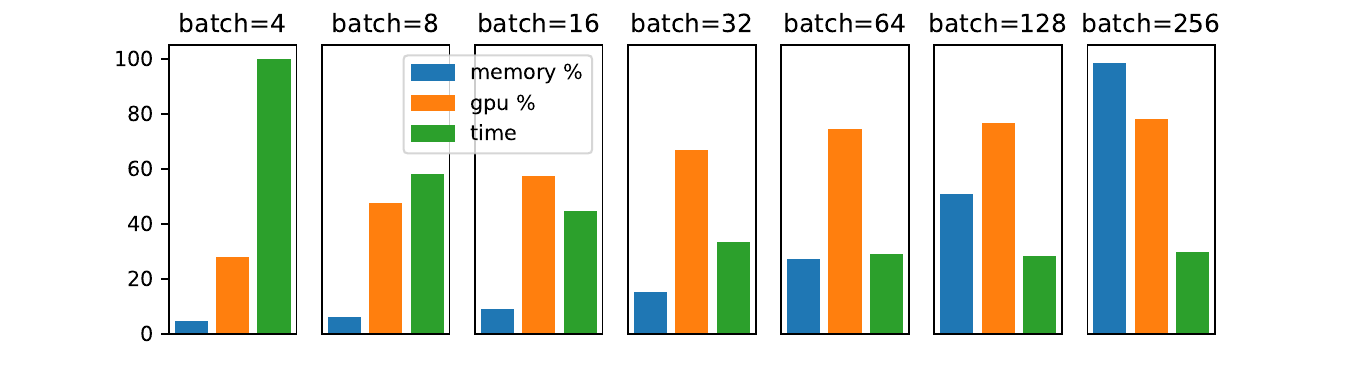}
    \caption{GPU usage metrics and computation time across batch sizes}
   \label{fig:usgmnist}
   \end{center}
\end{figure}

Figure \ref{fig:usgmnist} shows the GPU usage metrics on \textsc{Mnist} across different batch sizes for one client. We observe that memory usage and GPU utilization increase with higher batch sizes, while computation time decreases.

\begin{figure}[h]
    \begin{center}
    \includegraphics[scale = .4]{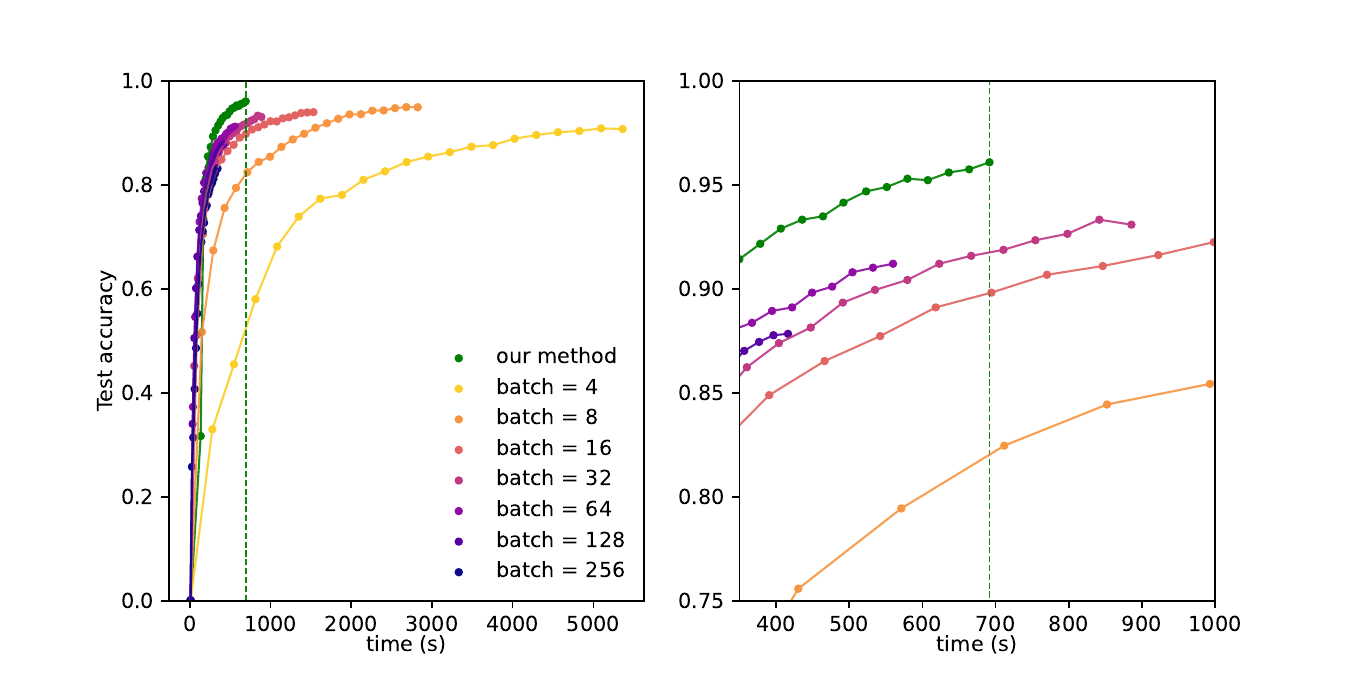}
    \caption{Accuracy with respect to wall-clock time on \textsc{Mnist}}
   \label{fig:accsmnist}
   \end{center}
\end{figure}

\begin{figure}[h]
    \begin{center}
    \includegraphics[scale = .4]{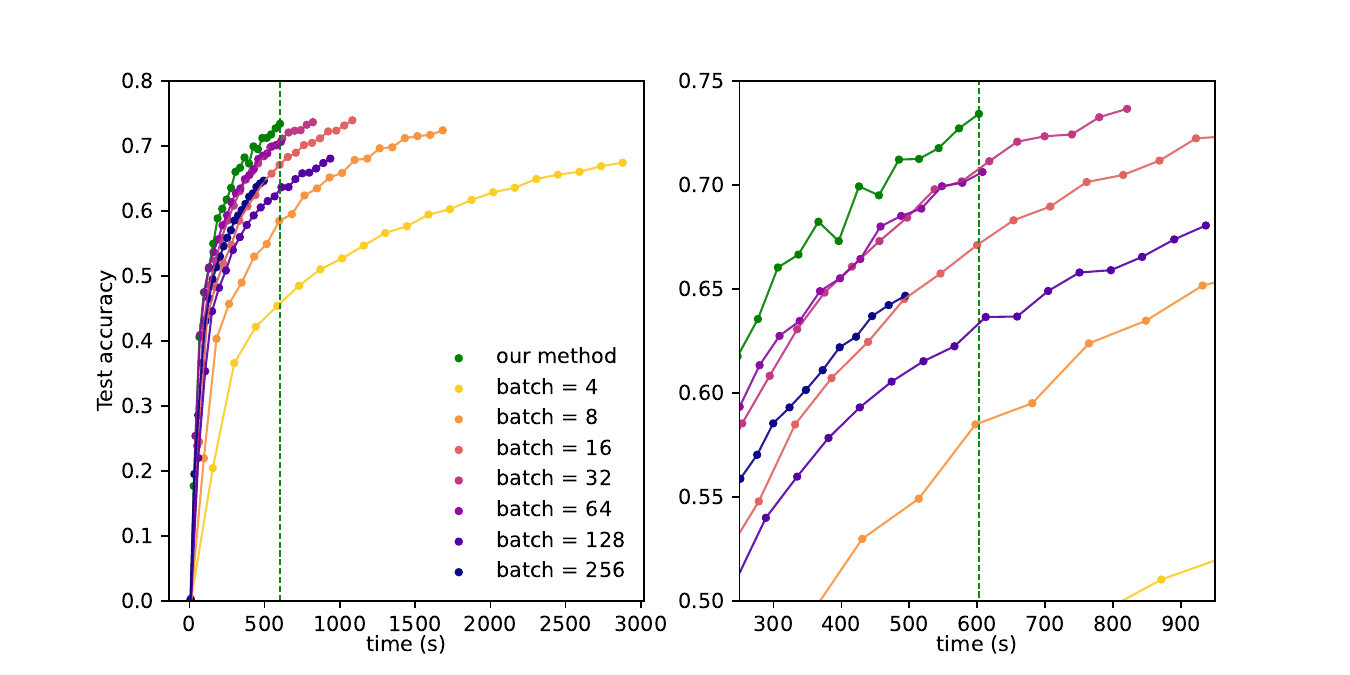}
    \caption{Accuracy with respect to wall-clock time on \textsc{Cifar10}}
   \label{fig:accscifar}
   \end{center}
\end{figure}

Figures \ref{fig:accsmnist} and \ref{fig:accscifar} show the accuracy over time of the global model on \textsc{Mnist} and \textsc{Cifar10}, respectively. In both cases, our experiments support previous findings that higher batch size values lead to lower computation time in Federated Learning. Furthermore, both figures show that our method, \textsc{RaSBa}, does not significantly increase the training time. With $f=0.5$ we observe that our method finds the maximal possible shared batch size in at most 3 rounds.

More details on the improved computation time are presented in Table \ref{table:speed}. Observations are twofold: First, we demonstrate the immense utility that larger batch sizes deliver. In fact, on \textsc{Mnist}, increasing the batch size from 4 to 256 decreased the time it takes the federation to complete by over $15\times$, while the increase yields a speed-up of $5.8\times$ on \textsc{Cifar10}. Secondly, we observe that our adaptive batch optimization method, allows us decrease the total computation time without much negative impact. Considering our method allowed a maximum batch size of 64, it performs closely to the federation with an initial mini-batch size fixed at 64.

In general, the performance of a federation using our method is close to optimal. Taking into account the possibility of failure when using a set batch size in Federated Learning, there is no reason not to use an adaptive batch size optimization to determine the maximal possible batch size and significantly speed up the Federated Learning process. We have shown that with only minimal loss in computation time, we are able to effectively determine a maximal value for the shared batch size in the first few rounds. 

\section{Conclusion and future work}

Our work highlights the problem of determining the maximum possible shared batch sizes in Federated Learning. Related research does not propose an automatic method of finding the limit, instead opting to determine the maximal value beforehand. As this value might surpass the system's capacity, previous work includes the potential exclusion of some participants or, in the worst case, the termination of the federation. By contrast, with just a minimal setback in overall computation time, our method ensures a smooth training process. It can be employed with any other Federated Learning method, be it a novel aggregation strategy or a client-specific training process.

Due to our method using the lowest common denominator as batch size, it can lead to some becoming stragglers. However, it ensures that all available clients are included in the training process, eliminating the risk of losing valuable data.

The domain in which our method can have the greatest impact is Federated Learning simulations. Instead of having to determine the limits of their hardware, researchers can utilize our optimal batch size search to maximize performance without having to test each possible batch size beforehand, saving considerable time. Especially when multiple participants are simulated in parallel on a single GPU, our proposed method considerably simplifies the task of optimizing batch size.
Future work should focus on improving our method when dealing with heterogeneous federations. In a scenario with multiple tiers of hardware, it might be interesting to group clients by the maximal batch sizes they report. Using a shared batch size per group, our method could lead to better scheduling and thus faster execution of federations.

\bibliographystyle{IEEEtran}
\bibliography{main}

\end{document}